\documentclass[journal]{IEEEtran}
\usepackage{cite}

\ifCLASSINFOpdf
  \usepackage[pdftex]{graphicx}
\else
\fi
\usepackage{amsmath}
\begin{document}
%
\title{Neural Network for NILM Based on Operational State Change Classification}
%
%
%

\author{Peng~Xiao,
Samuel~Cheng$^\ast$,~\IEEEmembership{Senior~Member,~IEEE}
\thanks{P. Xiao is with the Department
of Computer Science and Technology, Tongji University, Shanghai,
 201804 China (e-mail: phd.xiaopeng@gmail.com).}
\thanks{S. Cheng is with the School of Electrical and Computer Engineering, University of Oklahoma, OK 74105, USA (email: samuel.cheng@ou.edu).}
\thanks{$^\ast$ Corresponding author.}
}
%
%

{
%



\maketitle

\begin{abstract}
Energy disaggregation in a non-intrusive way estimates appliance level electricity consumption from a single meter that measures the whole home electricity demand. Recently, with the ongoing increment of energy data, there are many data-driven deep learning architectures being applied to solve the non-intrusive energy disaggregation problem. However, most proposed methods try to estimate the on-off state or the power consumption of appliance, which need not only large amount of parameters, but also hyper-parameter optimization prior to training and even preprocessing of energy data for a specified appliance. In this paper, instead of estimating on-off state or power consumption, we adapt a Deep Neural Network (DNN) to estimate the operational state change on appliance with single aggregate data. Our proposed solution is more feasible across various appliances and lower complexity comparing to previous methods. Through the simulated experiments in low-rate REDD dataset, we compare our proposed solution with two benchmark methods, Hidden Markov Model-based and Graph Signal Processing-based approaches, and a Recurrent Neural Network (RNN) featuring Gated Recurrent Units (GRU) architecture which estimates small window of aggregate data. All the results show the competitive performance of our proposed solution.
\end{abstract}


%
\IEEEpeerreviewmaketitle

\section{Introduction}
%
%
%
%

\IEEEPARstart 
{N}{on} intrusive load monitoring (NILM)\cite{Hart1992Nonintrusive} represents a pure computation
technique which identifies and extracts the power consumption of
individual appliances from measurements of the aggregate power usage for the entire home. 
Comparing to using appliance-level energy monitors, NILM minimizes maintenance and installation costs of sensors.
A significant application of NILM is to produce the individual electricity bills using readings from a single smart meter, which can deepen energy feedback leading to more efficient use of appliances and reduce the energy consumption.

In the last decade, with the smart energy meters 
have been being deployed in many countries, there is a growing number of datasets developed specifically for this research field. Thus a wide
variety of artificial intelligence and machine learning techniques applied to this problem, such as Hidden Markov Model (HMM) and its
variants\cite{Parson2012Non}, \cite{Zoha2012Non}, \cite{Kim2012Unsupervised}, \cite{Egarter2014PALDi}, decision tree\cite{Jing2014Non}, graph signal processing\cite{He2018Non}, support vector machines\cite{Lin2010Applying}, and non-negative tensor factorization\cite{Figueiredo2014Electrical}.

More recently, with the current breakthrough of deep neural networks in image classification\cite{krizhevsky2012imagenet}, speech recognition\cite{graves2014towards}, machine translation\cite{sutskever2014sequence}, 
DNN have regained their interests in addressing the disaggregation problem. Mauch and Yang\cite{mauch2015new} exploited a generic two-layer bidirectional
Recurrent Neural Network architecture featuring Long Short Term Memory (LSTM)\cite{hochreiter1997long} units in extracting single appliance profiles. In a latter work, Mauch
and Yang\cite{mauch2016novel} continued using a combination of discriminative and generative models in a two-stage eventless extraction of appliance profiles.  
Kelly and Knottenbelt\cite{kelly2015neural} evaluated and compared three neural network architectures, a RNN architecture with LSTM units similar to \cite{mauch2015new}, a de-noising Auto-Encoder(dAE)\cite{vincent2010stacked}, and a regression-based disaggregator which estimates the main key points of an activation cycle of the target appliance. Nascimento \cite{do2016applications} applied three deep neural network architectures, a basic convolutional dAE, a RNN, and a ResNet-based model\cite{he2016deep}, by introducing several improvements such as redefining the loss function, exploiting batch normalization \cite{ioffe2015batch}, and applying residual connections\cite{he2016deep}. All these
papers use synthetic data by summing all sub-meters, which limits the amount of noise as appliances not sub-metered would be excluded.
He and Chai\cite{he2016empirical} applied two architectures, a convolutional dAE and an RNN, with different kernel sizes applied to parallel convolutional layers.  Zhang et al.\cite{zhang2018sequence} simplified the objective of the dAE architecture in\cite{kelly2015neural} to predict a single time instance of the target appliance profile for a given window of the aggregated power. Murray et al.\cite{murray2018transferability} applied two architectures, a Convolutional Neural Network (CNN) and a bidirectional RNN featuring Gated Recurrent Units\cite{chung2014empirical}, to both estimates the state and the average consumption of targeted appliances. However, among these
works, each disaggregation window length (and consequently the width of subsequent layers) depends on the specified appliance being monitored, which
is not feasible in practical application. Said and Yang\cite{barsim2018feasibility} applied a deep fully convolutional neural networks to estimate a variety of load categories, but the model contains large number (i.e., 44) layers which is costly computational.   
Additionally, Lange et al.\cite{lange2016bolt} adopted a deep neural network with constrained binary and linear activation units
in the last two layers that estimate the on-off activation vector of each load, however it was applied on very high frequency current measurements. 

Review all above works, each deep neural network is designed to disaggregate a window of aggregate power and to estimate the on-off state or the power consumption of each target appliance. 
In the training process, each neural network needs to use large fraction or even entire power data of several houses, and the test usually carried only in one house (and some with data preprocessing), which is lack of persuasiveness. In fact, except estimate the state or the power consumption of appliance, there is an emerging field of NILM in estimating the operational state change of each appliance. In NILM, operational state change is defined as substantial statistical change in the aggregate power measurement occurs that indicates that one or more appliances have been switched on or off, or change their operational state. After such operational state change are identified, it will be classified into the predefined appliance categories. Through estimating the operational sate change of appliance, the state of appliance at each time instance can be determined by the sign of the power variation. Recently, the operational state change estimation in NILM gains some breakthroughs by using  GSP\cite{He2018Non}, but there is few research that applied deep neural network to estimate the operational state change of appliance. In this work, we apply a DNN to estimate the operational state change of each appliance with single aggregate data. And show competitive performance through experiments comparing to other two benchmark methods and a RNN architecture.
The main contributions of this paper are: 

(1) We show the neural network can perform well in estimating the operational change of appliance with a simple architecture. 

(2) Our proposed solution represents a significant reduction in complexity compared to previous works\cite{mauch2015new}, \cite{mauch2016novel}, \cite{kelly2015neural}, \cite{do2016applications}, \cite{he2016empirical}, \cite{zhang2018sequence}, \cite{murray2018transferability}, \cite{barsim2018feasibility}, \cite{lange2016bolt} in estimating the state or the
power consumption of appliance.

(3) Unlike previous deep architecture which need to input a window of aggregate power, our proposed solution can estimate the results at each time instance, which is more meaningful and feasibility across a variety of appliances. 

(4) The training process only use small percentage of data, and the test
is performed on the rest of raw dataset without any processing, which is 
occurs in many previous works, like balanced test data\cite{murray2018transferability} and sythesis data\cite{mauch2015new}, \cite{mauch2016novel}, \cite{kelly2015neural}, \cite{do2016applications}.    

The rest of the paper is organized as follows. In section \ref{section 2}, we describe the NILM task and the operational state change. In section \ref{section 3}, we describe the architecture of the proposed neural network. Section \ref{section 4} shows the experiment. Conclusion and future work are discussed in section \ref{section 5}.
 

\section{Load Disaggregation Based on Operational State Change}
\label{section 2}
In this section, we will introduce the load disaggregation task and the operational state change of each appliance.

\subsection{Load Disaggregation}
Let $\mathcal{A}$ be the set of all known appliances in a house and $P(t_i)$ be the
aggregate power of the entire house measured at time $t_i$. Without
loss of generality, in the following, we denote $P(t_i)$ as $P(t_i) = P_i \geq 0$.
Let $P^a_j \geq 0$ be the power load of appliance $a \in \mathcal{A}$ at time instance
$t_j$.  Then,
\begin{equation}
P_i = \sum_{a=1}^{|A|} P^a_i + n_i,
\end{equation}
where $n_j$  is the measurement noise that not submetered. The disaggregation task is for $i=1,...,T$   and  $a \epsilon \mathcal{A}$, given the $P_i$  to estimate the  $P^a_i$.

\subsection{Operational State Change}
Naturally, $\Delta P_i = P_{i+1} - P_i, i=1,...,T$ and $\Delta P^a_i = P^a_{i+1} - P^a_i, i=1,...,T$
respectively correspond to the variation of the aggregate power and appliance $a$ power
measured at time $t_i$. The classification labels of each appliance $a$ at time $t_i$ is denoted
as $s^a_i$. The $s^a_i$ is defined
as following:
\begin{equation}
 s^a_i = 
 \begin{cases}
  1, & \text{for $|\Delta p^a_i| \geq Thr_a$} \\
  0, & \text{for $|\Delta p^a_i| < Thr_a$},
 \end{cases}
\end{equation}
where $Thr_a \geq 0$ is a power threshold for appliance $a$. Like in previous literature\cite{He2018Non}, $Thr_a$ is set to half of difference between mean values of appliance $a$'s adjacent states, which is observed through training process. When $s^a_i$ is set to 1, it means appliance $a$ changed its operational state
 (e.g., switched on/off) at time $t_i$; When $s^a_i$ is set to 0, it means appliance $a$ didn't change its operational state at time $t_i$. $N_{pos}$ and $N_{neg}$ denote the number of the positive samples ($s^a_i = 1$) and negative samples ($s^a_i = 0$) respectively.

\subsection{Detect Operational State Change}
As can be seen in Fig. \ref{Figure 1}, at each time sample $t_i$, given the $\Delta P_i$ (i.e., the 
rising edge or the falling edge or the horizontal line), we estimate the $s^a_i$ which show whether appliance $a$ changed its operation state at time $t_i$.
\begin{figure}
\centering
\includegraphics[scale=0.5]{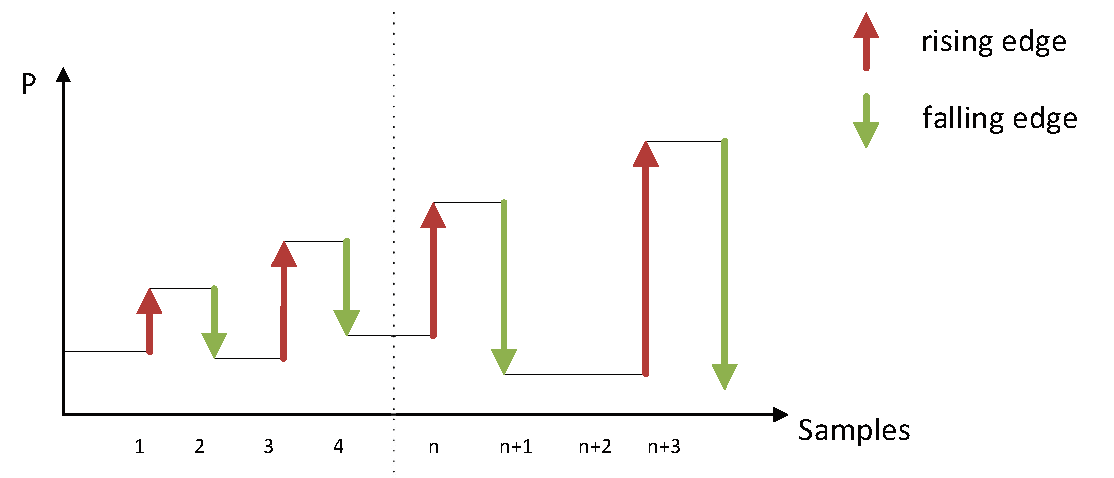}
\caption{An example of the operational state change.}
\label{Figure 1}
\end{figure}

\section{Proposed Network Architecture}
\label{section 3}
In the previous subsection, we have framed the operational state change detection task, now we describe how we adapted neural network to solve it.

Fig. \ref{Figure 2} show the detail architecture of the neural network. We propose a deep neural network (DNN) to calculate the classification labels for each appliance $a$.  The model consists of five layers reaching 1200 trainable parameters. Specifically, given the  aggregate power variation of the entire house $\Delta P_i$ at time $t_i$, firstly we initialize the input,
\begin{gather}
 x^{(0)} = |\Delta P_i|, 
\end{gather}
by using the absolute value of the aggregate power variation, then input it through the neural network. Each layer includes a sequence of elementary operations shown in the figure and briefly introduced in the sequel.

\textbf{Fully Connected:} the linear operation of each layer defined as a weighted multiplication and adding a bias:
\begin{equation}
f(x) = W^{(d)}x^{(d-1)} + b^{(d)},
\end{equation}
where $d$ denotes the layer number, $1 \leq d \leq D$, $W^{d}$ and $b^{d}$ denote the weight and bias of this layer respectively.

\textbf{Tanh:} is the non-linear activation function defined as:
\begin{equation}
f(x) = \frac{sinh(x)}{consh(x)} = \frac{e^x - e^{-x}}{e^x + e^{-x}},
\end{equation}
the non-linear activation is the component which convert the linear operation to non-linear.

\textbf{Batch Normalization:} is a composition of two affine transformation applied to the output of each layer based on mini-batch statistic:
\begin{equation}
f(x) = \gamma \widehat{x} + \beta = \gamma \frac{x - \mu_{\beta}}{\sigma_\beta} + \beta,
\end{equation}
where $x$  is the original output of a unit, $\mu_\beta$  and $\sigma_\beta$  are the sample mean and standard variation of all outputs of this neuron over the mini-batch  $B$,  $\gamma$ and $\beta$  are two learnable parameters. The batch normalization is very important in neural network. When we apply it to the neural network, the performance is improved a lot.

\textbf{Softmax:} is an activation function applied to the output of last layer in the model:
\begin{equation}
f(x)_i = \frac{e^{x_i}}{\sum^K_{k=1} e^{x_k}} \text{ for } i = 1, ..., K,
\end{equation}
which calculate the normalized probability distribution of a output vector, and $K$  denotes the number of the class (in this paper,  $K=2$).

 After the features are extracted through the neural network, it will be passed to the:  
\begin{gather}
\mathit{ p_i}^a = softmax(x^{(D)}),
\end{gather}
to calculate a binary likelihood probabilities $\mathit{p_i}^a =(p_0, p_1)$, which represent the probabilities of whether the appliance changed it's state or not respectively.

\begin{figure}
\centering
\includegraphics[scale=0.5]{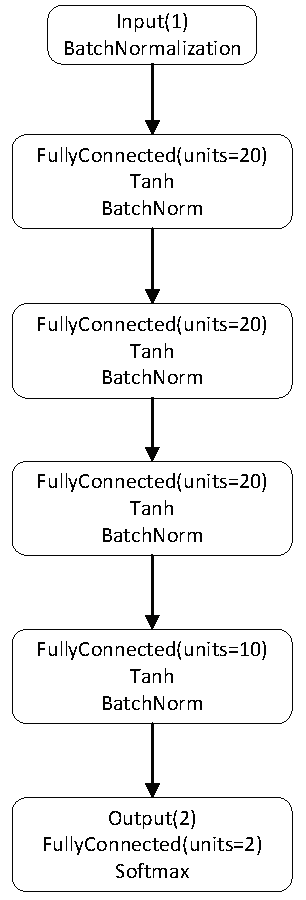}
\caption{The detail architecture of the proposed model.}
\label{Figure 2}
\end{figure}

\subsection{Loss Function}
The cost for our neural network training is the negative log-likelihood:
\begin{gather}
J({\mathbf{p}^a}, \mathbf{{s}^a}) = \sum_{i} -\log \mathit{p}^a_i[s^a_i],
\end{gather}
where $\left[ \cdot \right]$ is the index operate.

\section{Experiment}
\label{section 4}

\subsection{Dataset}
We evaluate the proposed model on the REDD dataset\cite{kolter2011redd} downsampled to 1 min resolution as in \cite{He2018Non}.

\textbf{REDD} dataset\cite{kolter2011redd} is a dataset for
energy disaggregation.  The dataset contains about half month power consumption from real
homes in US, for the whole house as well as for each individual circuit in
the house (labeled by the main type of appliance on that circuit). The
main types of the appliances are: Dishwasher (DW), Refrigerator (REFR), 
Microwave (MW), Kitchen outlet (KO), Stove (ST), air-conditioning (AC), Electronics (EL), Wash Dryer (WD). In this experiment, we use three 
houses energy data: House 1, House 2, House 6. A week of of the REDD House 1 data can be seen in Fig. \ref{Figure 3}.
\begin{figure}
\centering
\includegraphics[scale=0.5]{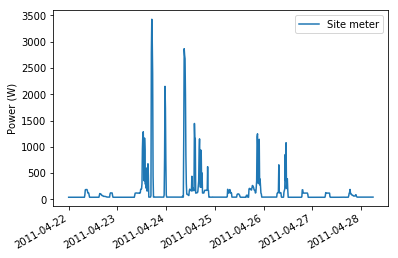}
\caption{A week power data of REDD House1.}
\label{Figure 3}
\end{figure}

\subsection{Training}
We train the proposed architecture using a small part data of House1, House2, House6 in REDD. The total samples of House1, House2 and House6 
are 25946, 19856, 17605 respectively. The training detail of three house are respectively shown in Table \ref{Table 1}, Table \ref{Table 2}, Table \ref{Table 3}
\begin{table}[h]
\caption{THE TRAINING DETIALS OF HOUSE 1}
 \label{Table 1}
\begin{center}
 \begin{tabular}{|c|ccccc|}
  \hline
  Appliance & REFR & MW & DW & KO & WD \\
  \hline
  Training Samples & 2000 & 2000 & 5000 & 2000 & 8000 \\ 
  Thresold(Watts) & 150 & 750 & 210 & 550 & 1300 \\ 
  \hline
 \end{tabular}
\end{center}
\end{table}

\begin{table}[h]
 \caption{THE TRAINING DETAILS OF HOUSE 2}
 \label{Table 2}
 \begin{center}
  \begin{tabular}{|c|cccc|}
   \hline
   Appliance & REFR & MW & KO & ST \\
   \hline
   Training Samples  & 2000 & 2000 & 2000 & 4920\\ 
   Thresold(Watts) & 85.5 & 920 & 528 & 204\\ 
   \hline
  \end{tabular}
 \end{center}
\end{table}

\begin{table}[h]
 \caption{THE TRAINING DETAILS OF HOUSE 6}
  \label{Table 3}
 \begin{center}
  \begin{tabular}{|c|ccccc|}
   \hline
  Appliance & REFR & AC & EL & KO & ST \\
  \hline
  Training Samples & 2000 & 2000 & 4640 & 3000 & 3445 \\ 
  Thresold(Watts) & 74.5 & 862  & 225 & 660 & 1700 \\ 
  \hline
  \end{tabular}
 \end{center}
\end{table}

\subsection{Augment positive samples}
In real life, many appliances are not commonly used, such as stove, wash dryer etc., which means the positive samples
is much less than the negative samples in the training set, and that will lead the model output a unbalance result (classify all the samples to 
negative). 

Let us denote $N_{pos}$ and $N_{neg}$ as the number of the positive samples ($s^a_i=1$) and negative samples ($s^a_i = 0$) in the
training set respectively.
In order to solve the unbalance issue, during training, we augment the positive samples
in the following way:
\begin{enumerate}
\item calculate the ratio between the negative samples and positive samples in the
training dataset, $\eta = N_{neg} / N_{pos}$;
\item determine a ratio of positive to negative $\alpha$, then calculate the positive
scaling factor $\sigma = \eta * \alpha$;
\item duplicate the positive samples by $\sigma$ times, and randomly insert them into the
original training dataset.   
\end{enumerate} 
The $\alpha$ in this paper is set to 1:8.

\subsection{Evaluation metrics}
The evaluation metrics used are precision (PR), recall (RE) and F-Measure ($F_M$)\cite{Delen2008Advanced} defined as:
\begin{gather}
 PR = TP / (TP + FP)\\
 RE = TP / (TP + FN)\\
 F_M = 2 * (PR * RE)/(PR + RE),
\end{gather}
where true positive (TP) is recorded when the state of the detected appliance was actually changed, false positive (FP) is recorded when the
state of the detected appliance was not changed, and false negative (FN) indicates that the state changed appliance was not detected.
Precision captures the correctness of detection, and the high Recall implies a higher percentage of appliance state changes are detected
correctly. 

\subsection{Setup}
We use PyTorch\cite{Paszke2017Automatic} to develop our model and optimize the model by  Adam\cite{Kingma2014Adam} optimizer with a base learning rate 1e-4, momentum 0.99.  The data 
is processed by the Numpy and Pandas. 

\subsection{Comparison with benchmarks}
We compare our model with two benchmark methods, HMM-based approach\cite{Parson2012Non} and GSP-based approach\cite{He2018Non}. The results are
shown in Table \ref{Table 4}, Table \ref{Table 5}, Table \ref{Table 6} for Houses 1, 2, 6
respectively.

\begin{table}[h]
 \caption{COMPARISION RESULTS OF THREE METHODS IN HOUSE 1}
 \label{Table 4}
\begin{center}
 \begin{tabular}{|c|ccccc|}
  \hline
  Appliance & REFR & MW & DW & KO & WD \\
  \hline
  $F_{M_{NN}}$ & 0.88 & 0.76 & 0.47 & 0.64 & 0.88 \\ 
  $F_{M_{GSP}}$ & 0.88 & 0.70 & 0.57 & 0.39 & 0.89 \\ 
  $F_{M_{HMM}}$ & 0.97 & 0.50 & 0.13 & 0 & 0 \\ 
  \hline
 \end{tabular}
\end{center}
\end{table}

\begin{table}[h]
 \caption{COMPARISION RESULTS OF THREE METHODS IN HOUSE 2}
 \label{Table 5}
 \begin{center}
  \begin{tabular}{|c|cccc|}
   \hline
   Appliance & REFR & MW & KO & ST \\
   \hline
   $F_{M_{NN}}$ & 0.85 & 0.97 & 0.91 & 0.83\\ 
   $F_{M_{GSP}}$ & 0.84 & 0.93 & 0.88 & 0.86\\ 
   $F_{M_{HMM}}$ & 0.90 & 0.47 & 0.68 & 0.21\\ 
   \hline
  \end{tabular}
 \end{center}
\end{table}

\begin{table}[h]
 \caption{COMPARISION RESULTS OF THREE METHODS IN HOUSE 6}
 \label{Table 6}
 \begin{center}
  \begin{tabular}{|c|ccccc|}
   \hline
  Appliance & REFR & AC & EL & KO & ST \\
  \hline
  $F_{M_{NN}}$ & 0.80 & 0.89 & 0.70 & 1 & 0.90 \\ 
  $F_{M_{GSP}}$ & 0.77 & 0.88 & 0.66 & 0.88 & 0.92 \\ 
  $F_{M_{HMM}}$ & 0.99 & 0.12 & 0.03 & 0 & 0 \\ 
  \hline
  \end{tabular}
 \end{center}
\end{table}

As can be seen from Tables \ref{Table 4}, Table \ref{Table 5} and Table \ref{Table 6}, our proposed solution
outperforms the other two methods in many cases, which shows 
the superiority of our method in estimating the operational state change of appliance.
Specifically, our method significantly outperforms the HMM-based method in all appliances except the refrigerator. 
This is mainly due to continuous and sole operation of refrigerator, hence there is large 
available data for learning and improving initial HMM model. The poor performance of HMM for other appliances can be attributed to
the short training period.
The proposed model shows better or similar performance to the GSP-based method, especially Kitchen Outlets in three houses. This is mainly
due to the large fluctuations during operation, thus the GSP-based method cannot accurately capture the appliance operation.

All the results for multi-state appliances (dishwasher in House 1) are generally worse for all three methods. This is due to the similarity between the 
refrigerator load and low-state of the dishwasher, so they are often `hidden' in the baseload and noise. And multi-state appliances are not
used frequently, thus it is more difficult to extract during the training phase.  

Additionally, the training details in Table \ref{Table 1}, Table \ref{Table 2} and Table \ref{Table 3} show the training condition differences between different kinds of appliances. Stove usually 
needs more training data, which is due to the fact that Stove normally has short operation time
and relative high power, thus the neural network needs more data to generate the probabilistic
models to capture the appliance operational state change. Electronics and
Wash Dryer  are not used often in common houses, so the model also needs more data to
learn the statistically pattern of these two appliances.

\subsection{Comparison with RNN}
We also compare our solution with a GRU-based RNN architecture. The GRU is a variant of the LSTM unit, especially designed for time series 
data. Comparing to LSTM, GRUs have fewer parameter and are more suited to online learning.  

We adapt the GRU-based RNN to estimate two time-step and three time-step window of aggregate data, and the results are shown in Table \ref{Table 7}, \ref{Table 8}, and \ref{Table 9} for REDD House 1, 2, 6 respectively. The architecture details of RNN can be seen in Fig. \ref{Figure 4}. For consistency, the number of layers and hidden features in RNN are same as those in DNN. And the training and test setup are also same as DNN setup.

\begin{figure}
\centering
\includegraphics[scale=0.5]{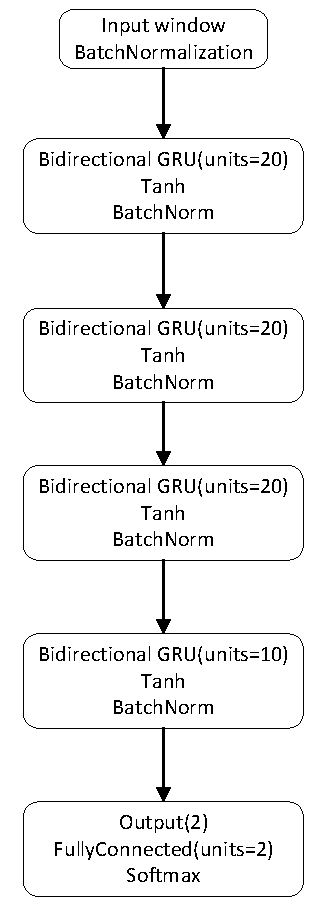}
\caption{Architecture of compared RNN model.}
\label{Figure 4}
\end{figure}

\begin{table}[h]
 \caption{COMPARISION RESULTS WITH RNN IN HOUSE 1}
 \label{Table 7}
\begin{center}
 \begin{tabular}{|c|ccccc|}
  \hline
  Appliance & REFR & MW & DW & KO & WD \\
  \hline
  $F_{M_{NN}}$ & 0.88 & 0.76 & 0.47 & 0.64 & 0.88 \\ 
  $F_{M_{RNN_2}}$ & 0.88 & 0.65 & 0.44 & 0.39 & 0.82 \\ 
  $F_{M_{RNN_3}}$ & 0.88 & 0.59 & 0.21 & 0.2 & 0.75 \\ 
  \hline
 \end{tabular}
\end{center}
\end{table}

\begin{table}[h]
 \caption{COMPARISION RESULTS WITH RNN IN HOUSE 2}
 \label{Table 8}
 \begin{center}
  \begin{tabular}{|c|cccc|}
   \hline
   Appliance & REFR & MW & KO & ST \\
   \hline
   $F_{M_{NN}}$ & 0.85 & 0.97 & 0.91 & 0.83\\ 
   $F_{M_{RNN_2}}$ & 0.87 & 0.84 & 0.70 & 0.4\\ 
   $F_{M_{RNN_3}}$ & 0.86 & 0.70 & 0.71 & 0.21\\ 
   \hline
  \end{tabular}
 \end{center}
\end{table}

\begin{table}[h]
 \caption{COMPARISION RESULTS WITH RNN IN HOUSE 6}
 \label{Table 9}
 \begin{center}
  \begin{tabular}{|c|ccccc|}
   \hline
  Appliance & REFR & AC & EL & KO & ST \\
  \hline
  $F_{M_{NN}}$ & 0.80 & 0.89 & 0.70 & 1 & 0.90 \\ 
  $F_{M_{RNN_2}}$ & 0.80 & 0.24 & 0.13 & 0.61 & 0.47 \\ 
  $F_{M_{RNN_3}}$ & 0.79 & 0.37 & 0.22 & 0.66 & 0.20 \\ 
  \hline
  \end{tabular}
 \end{center}
\end{table}
As can be seen in Tables \ref{Table 7}, \ref{Table 8}, \ref{Table 9}, without any other preprocessing (like synthesizing data or balancing data), the results of the GRU-based RNN architecture in estimating small window aggregate data are much worse than DNN in estimating a single aggregate data in most cases. This is mainly because of the 
low correlation between the aggregate sequence in operational state change. Intuitively, it's hard to estimate whether the appliance changes 
its operational state according to the situation in surrounding time. And in many types of appliance, we have many more negative samples than positive samples, which also limits the performance of RNN in estimating the operational state change.

\section{Conclusion}
\label{section 5}
This paper adapts a simple DNN to detect the operational state change of appliance in the NILM task. 
Comparing to previous deep learning architecture which estimates the on-off state or power consumption, the proposed
solution is low-complexity and more feasibility across variety kinds of appliance. 
Through the simulated 
experiments from three real houses in the REDD dataset comparing to other two benchmarks and a RNN architecture, we prove the 
neural network's competitive performance in estimating the operational state change of appliance. 
Unlike previous deep learning architectures which are trained on large part or even entire power data of several houses, our proposed method only needs a small percentage of one house data to achieve
a competitive performance. So the results also indicate the statistically
regularity of appliance operational state change in one house, which reveal the potential of 
neural network in detecting the
operational state change in NILM task.

In the future work, we will try to use the time information in a more reasonable way to improve the performance on more appliances,  and we will also try to disaggregate several or all appliances at once.


%



%



\section*{Acknowledgment}
This project has received funding from the European Union’s Horizon 2020 research and innovation programme under the Marie Sklodowska-Curie grant agreement no. 734331 and the Fundamental Research Funds for the Central Universities no. 0800219369.


\ifCLASSOPTIONcaptionsoff
  \newpage
\fi



%


\bibliographystyle{IEEEtran}
\bibliography{1,2,3}
\end{document}